\documentclass[10pt,a4paper]{article}
\usepackage{framed,multirow}

\usepackage{amssymb}
\usepackage{latexsym}

\usepackage{url}
\usepackage{xcolor}
\definecolor{newcolor}{rgb}{.8,.349,.1}

\usepackage{amsmath}
\usepackage{amsfonts}
\usepackage{amssymb}
\usepackage{color}
\usepackage{tikz}
\usepackage{booktabs}
\usepackage{adjustbox}
\usepackage{hyperref}

\usepackage[T1]{fontenc}
\usepackage[utf8]{inputenc}
\usepackage{authblk}

\title{
	Supervised Infinite Feature Selection
}
\author[1]{Sadegh Eskandari A\thanks{eskandari@guilan.ac.ir}}
\author[2]{Emre Akbas B\thanks{emre@ceng.metu.edu.tr}}
\affil[1]{Department of Computer Science, University of Guilan, Rasht, Iran}
\affil[2]{Department of Computer Engineering, Middle East Technical Unviersity, Ankara 06800, Turkey}

\begin{document}
\maketitle

\begin{abstract}
In this paper, we present a new feature selection method that is 
     suitable for both unsupervised and supervised problems. We build upon the
     recently proposed Infinite Feature Selection (IFS) method where feature subsets
     of all sizes (including infinity) are considered.  We extend IFS in two
     ways. First, we propose a supervised version of it. Second, we propose new
     ways of forming the feature adjacency matrix that perform better for
     unsupervised problems. We extensively evaluate our methods on many
     benchmark datasets, including large image-classification datasets (PASCAL
     VOC), and  show that our methods outperform both the IFS and
     the widely used ``minimum-redundancy maximum-relevancy (mRMR)'' feature selection algorithm. 
\end{abstract}

%\linenumbers

%% main text
\section{Introduction and Related Work}\label{Intro}
In many practical machine learning and classification tasks, we
encounter a very large feature space with thousands of irrelevant and/or
redundant features. Presence of such features causes high computational
complexity, poor generalization performance and decreased learning accuracy
\cite{Eskandari16a,Guyon03}.
The task of feature selection is to identify a small subset of most important,
i.e. representative and discriminative, features.
Many feature selection algorithms have been proposed  in the last three decades
(e.g. \cite{Guyon03,Kohavi97,Saeys07,Canedo13}). Among them, \emph{filters} have
generated much interest, because they are simple, fast and  not biased to any
special learner.  In these methods, each candidate feature subset is evaluated
independent of the final learner, based on a diverse set of evaluation measures
including mutual information \cite{Peng05,Brown12}, consistency \cite{Dash03},
significance \cite{Koller96,Wu13}, etc.

Most filter methods rely on the concept of feature relevance
\cite{Brown12,Koller96,Yu04}. For a given learning task, a feature can be in one
of the following three disjoint categories: strongly relevant, weakly relevant
and irrelevant. Strongly relevant features contain information that is not
present in any subset of other features and therefore they are always necessary
for the underlying task. Weakly relevant features contains information which is
already present in a subset of strongly  or irrelevant features. These features
can be unnecessary (redundant) or necessary (non-redundant) with certain
conditions. Irrelevant features contain no useful information and are not
necessary at all. An ideal feature selection algorithm should eliminate all the
irrelevant features and weakly redundant features. However, constructing such an
algorithm is computationally infeasible, as it requires to check exponentially
many combinations of features to ascertain weak relevancy. Therefore, several
heuristics are proposed in the literature, which consider limited combination
sizes \cite{Lewis92,Battiti94,Peng05,Fleuret04,Yu04,Meyer06,Wu13}.

Recently, an interesting filter method called ``infinite feature selection"
(IFS) was proposed by \cite{Roffo15}. This method ranks features
based on path integrals and the centrality concept on a feature adjacency graph. The most
appealing characteristics of this approach are 1) all possible subsets of
features are considered in evaluating the rank of a given feature and 2) it is
extremely efficient, as it converts the feature ranking problem to simply
calculating the geometric series of an adjacency matrix.
%For example, for a dataset with 1K instances and 10K features, it takes less than 20 seconds to rank all the features, on a regular PC.
Although it outperforms most of the state-of-the-art feature selection methods
in image classification and gene expression problems, the algorithm suffers from
two important deficiencies. Firstly,  it is an unsupervised feature selection
algorithm, i.e. it does not use the provided labels in a supervised learning
problem. Secondly, its feature redundancy measure is not able to capture
complex non-linear dependencies.

In this paper, we improve the IFS method in two ways. First, we propose a
method to form the feature adjacency matrix for supervised problems. Second, we
propose alternative ways of forming the adjacency matrix for unsupervised
scenarios. In our experiments, we extensively compare our new methods with IFS
and other popular feature selection methods. We show that our proposed methods
outperform IFS on many different benchmark datasets and large
image-classification datasets (PASCAL VOC 2007 and 2012) as well\footnote{Leaderboard
snapshot taken in December 2016:\\
\url{http://user.ceng.metu.edu.tr/~emre/resources/SIFS_PASCAL_result.png}.\\ Our
submission is named "SE."
Anonymous results link:
\url{http://host.robots.ox.ac.uk:8080/anonymous/MV5IFE.html}.  Live
leaderboard:
\url{http://host.robots.ox.ac.uk:8080/leaderboard/displaylb.php?challengeid=11&compid=1}}.
Source code of our methods will be released upon acceptance of the paper.

The remainder of the paper is organized as follows: Section \ref{section:inf-FS}
presents the general idea  behind IFS. Section \ref{section:AM} discusses the feature adjacency matrix  along with our proposals for proper construction of this matrix in supervised and unsupervised feature selection. Section \ref{section:ER} reports experimental results and Section \ref{section:Conclusions} concludes the paper.

\section{Infinite Feature Selection}\label{section:inf-FS}

In this section, we review the general idea behind the IFS algorithm as proposed
by \cite{Roffo15}, for completeness. Given a dataset with $m$ features $\{f_1,f_2,\dots, f_m\}$, an
undirected complete weighted graph $G=\left(V,E,e\right)$ can be constructed such that $V=\{f_i|f_i \in F \}$
represents the vertices, $E=\{\{f_i,f_j\}|f_i,f_j \in F \wedge i\neq j\}$
represents the edges and $e: E \rightarrow  \mathbb{R}$ is a function calculating
the pairwise energies between features.  $G$ can be represented using an adjacency matrix $A$, such that $a_{ij}=e(\{f_if_j\})$.
Let $P^{l}_{i,j}$ be the set of all paths of length $l$ (including paths with
cycles) between nodes $i$ and $j$, and let $A^l$ denote the power iteration of
matrix $A$. An initial idea for feature selection could be choosing an
appropriate length $l$, then calculating energy scores, $s_l(i)$, for each
feature $f_i$ as:
\begin{equation}\label{eq:energy_score}
	s_l\left(i\right)=\sum_{j\in V}\sum_{p\in P^{l}_{i,j}} \prod_{k=0}^{l-1}a_{v_k,v_{k+1}}=\sum_{j\in V}A^l \left(i,j\right),
\end{equation}
and finally taking a subset of features with maximum energy value. However, this
idea has two major drawbacks; first, cycles can have high impact
in calculating the scores and second, computation of $A^l$ is of order
$O\left(n^4\right)$, which is impractical when the number of features is large.
The main contribution by \cite{Roffo15} is to address the deficiencies by
expanding the path length to infinity and summing over all path lengths. By
extending the path length to infinity, the probability of being part of a cycle
is uniform for all the features so the cycle effect is somewhat normalized.
Therefore, a new energy score for each feature $f_i$, considering all path
lengths including infinity,   can be calculated as:

\begin{equation}
    s\left(i\right)=\left[ \left( \left(\sum_{l=0}^{\infty}A^l
    \right)-\mathrm{I}\right) \overline{\textbf{1}} \right]_i
\end{equation}
where I is the identity matrix and $\overline{\textbf{1}}$ is a column vector of ones.

In matrix algebra,  $\sum_{k=0}^{\infty}X^l$ is called the geometric series of
matrix $X$. This series converges to $\left(\mathrm{I}-X\right)^{-1}$ if and only if $\rho(X)<1$, where $\rho(X)$ is the maximum
magnitude of the eigenvalues of $X$. For any matrix $X$, it can be shown that $\rho(rX)<1$ if and only if $0<r<\frac{1}{\rho(X)}$. Using this property, the regularized energy score for each feature $f_i$ can be defined as
\begin{equation}
            \begin{split}
                s'(i) & = \left[ \left( \left(\sum_{l=0}^{\infty}r^l A^l
                \right)-\mathrm{I}\right) \overline{\textbf{1}} \right]_i \\
                & = \left[ \left( \left( \mathrm{I}-rA\right)^{-1}-\mathrm{I}\right)
                \overline{\textbf{1}} \right]_i.
            \end{split}	
\end{equation}
Therefore, the computation of power iterations of matrix $A$ in
Eq.\ref{eq:energy_score}, is reduced to computing $\left( \left(
\mathrm{I}-rA\right)^{-1}-\mathrm{I}\right)$, with a complexity of $O\left(n^{2.37}\right)$.

\section{Forming the Adjacency Matrix} \label{section:AM}
As explained in Section \ref{section:inf-FS}, the IFS algorithm uses the
adjacency matrix $A$ to compute ranking scores for given feature distributions.
Therefore, the formation of the matrix can be considered as the most important
task in the approach. In this section, we  propose new  ways of  constructing
the matrix $A$ both  for supervised and unsupervised feature selection scenarios.

\subsection{Unsupervised Feature Selection (mIFS)}\label{section:unsupervised_feature_selection}
Defining feature relevance in unsupervised  learning is a big challenge, because
we do not know a-priori what type of patterns to look for or which error metric
to use. Furthermore, these two aspects often depend on the dataset used. However, one can analyse the features in terms of redundancy and dispersion.

If a certain feature has zero dispersion (i.e. variance) over the examples in
the dataset, then that feature does not have any information and can be
discarded. For a feature with non-zero dispersion, although we can not
definitively relate its relevance to its dispersion magnitude, it has been shown that
using dispersion measures improves the performance \cite{Roffo15,Guyon03}. Let
$STD_f$ be the standard deviation of feature $f$. Our experiments also show that
keeping features that have large standard deviation, i.e. $STD_f$, improves the classification accuracy.

 The other measure we use in unsupervised feature selection is redundancy.
 Unlike relevance, redundancy is a well-defined problem in unsupervised learning
 and can be expressed in terms of dependency. For example, when the dependency
 among two disjoint feature subsets is large, one of them could be
 considered as redundant. \cite{Roffo15} used the Spearman's rank
 correlation coefficient  as a measure of redundancy of a feature. However, this
 measure is not able to individuate complex non-linear dependencies between
 features (e.g. non-monotonic non-linear dependencies). Our experiments show
 that using a mutual information-based measure for redundancy yields  better results in terms
 of classification accuracy. This is probably due to the fact that mutual information takes into account any kind of
 dependency (both linear and non-linear) between random variables
 \cite{Dionisio04}. For a given feature set $F$ and a feature $f\in F$, we
 define this measure as:
\begin{equation}
    RDN_f=\frac{1}{|F|-1}\sum_{f'\in F-\{f\}}\mathrm{MI}(f',f)
\end{equation}
where, $\mathrm{MI}(X,Y)$ is the mutual information between two random variables $X$ and $Y$, and is defined as
\begin{equation}
    \mathrm{MI}(X,Y)=\int_{\mathcal{X}} \int_{\mathcal{Y}}p\left(X=x,Y=y\right)
    \log \left(   \frac{p\left(
    X=x,Y=y\right)}{p\left(X=x\right)p\left(Y=y\right)}  \right).
\end{equation}

Overall, we propose the following adjacency matrix to be used in unsupervised feature selection scenarios:

\begin{equation}\label{eq:unsupervised_matrix}
\begin{split}
 a_{ij}  &= \alpha \left(\max \left(STD_{f_i},STD_{f_j}\right)\right) \\
 &+ (1-\alpha)\left( 1-\min\left(RDN_{f_i},RDN_{f_j} \right) \right),
\end{split}               
\end{equation}
where $\alpha \in [0,1]$ is a loading coefficient that controls the relative
importance of relevance vs. redundancy. We name this way of constructing $A$ as
the modified infinite feature selection or mIFS, for short.

\subsection{Supervised Feature Selection (SIFS)}\label{section:supervised_feature_selection}
In supervised machine learning, the goal is to learn a general form of an
unknown mapping from a feature vector $f$ to a target variable $Y$. Therefore,
the relevance of features can be expressed in terms of the $Y$-related
information they have. Mutual information would be a proper measure to capture
this relevance.

We augment the supervised relevancy measure with an unsupervised redundancy
measure.  Although mutual information based redundancy yields
good accuracy for unsupervised scenarios, our experiments show that when it is
combined with the same measure for relevance analysis, the accuracy deteriorates
significantly. Therefore, for supervised feature selection, we propose to use
Spearman's rank correlation based redundancy. Specifically, we propose the following adjacency matrix:
\begin{equation}\label{eq:supervised_matrix}
\begin{split}
a_{ij}  &= \alpha \left(\max \left(\mathrm{MI}(f_i,Y),\mathrm{MI}(f_j,Y)\right)\right)\\ 
&+ (1-\alpha)\left( 1- |SPR \left(f_i,f_j\right)|\right),
\end{split}                
\end{equation}
where $SPR(X,Y)$ is the Spearman’s rank correlation coefficient. We name this
method as supervised infinite feature selection or SIFS, for short.

\section{Experimental Results} \label{section:ER}
We conducted three sets of experiments. First, as preliminary experiments, we
explored  the effects of different ways of constructing the adjacency matrix on
the classification performance of IFS and SIFS. Next, we compared the
classification performances of the IFS with original settings, IFS with the
adjacency matrix proposed in Eq.\ref{eq:unsupervised_matrix} (i.e. mIFS), SIFS with the
adjacency matrix proposed in Eq.\ref{eq:supervised_matrix} and the well-known
minimum-redundancy maximum-relevancy (mRMR) algorithm proposed by \cite{Peng05}.
Finally, we focused on the image classification problem where we used SIFS to select features from the state-of-the-art convolutional neural networks (CNN).

\begin{table*}
\caption{Summary of the high dimensional benchmark datasets together
with their main challenges and the state of the art (SoA)
performances. The star(*) in the last column indicates that our methods
achived a new SoA for the corresponding dataset.}
\label{table:benchmark-datasets}
\centering
\footnotesize
\begin{tabular}{lcccccl}
\hline
%\multicolumn{2}{c}{Item} \\
%\cline{1-2}
\textbf{dataset}                    &\textbf{ \#feat.}       & \textbf{\#classes}  & \textbf{\#samples}  &   \textbf{few train}  & \textbf{noise} &  \textbf{SoA} \\
\hline
USPS \cite{Chapelle06}            &            241         &             2
    &            1.5K     &                       &                &96.6\%
    \cite{Maji09}\\
GINA \cite{WCCI06}                &            970         &             2
    &            3153     &                       &                &99.7\%
    \cite{Guyon08}          \\
Gissete \cite{Nips03}             &            5K          &             2
    &            7K       &                       &                &99.9\%
    \cite{Guyon07}          \\
\hline
Colon \cite{Alon99}               &            2K
&             2       &            62       &
\checkmark    &   \checkmark   &89.6\%
\cite{Lovato12}~~*          \\
Lung181 \cite{Gordon02}           &            12533
&             2       &            181      &
\checkmark     &  \checkmark    &99.8\% \cite{Roffo15}~~*          \\
DLBCL   \cite{Shipp02}            &            7129
&             2       &            77       &
\checkmark      &   \checkmark   &98.3\% \cite{Roffo15}~~*          \\
Prostate \cite{Singh02}           &            6033        &             2
    &            102      &       \checkmark      &   \checkmark   &99.94\%
    \cite{Diaz06}          \\
Arcene \cite{Nips03}              &            10K         &             2
    &            200      &                       &                &99.93\%
    \cite{Neal06}          \\
REGED0 \cite{WCCI08}              &            999         &             2
    &            20.5K    &       \checkmark      &                &100\% \cite{Chang08}          \\
MARTI0 \cite{WCCI08}              &            999         &             2
    &            20.5K    &       \checkmark      &                &99.94\%
    \cite{Cawley08}          \\
\hline
Madelon  \cite{Nips03}            &            500         &             2
    &            2.6K     &                       &                &98.0\%
    \cite{Guyon07}          \\
\hline
Sido0  \cite{WCCI08}              &            4932
&             2       &            22678    &
&                &94.7\% \cite{Guyon08b}          \\
\hline
VOC 2007 \cite{pascal-voc-2007}   &\textit{not specified}
&             20      &            9963     &
&   \checkmark   &83.5\% \cite{Roffo15}          \\
VOC 2012 \cite{pascal-voc-2012}   &\textit{not specified}
&             20      &            22531    &
&   \checkmark   &85.4\% \cite{pascal-voc-2012}        \\

\hline
\end{tabular}
\end{table*}

Table \ref{table:benchmark-datasets} summarizes the 14 high-dimensional
benchmark datasets that we used in our experiments. These benchmarks include
handwritten character recognition (USPS, GINA and Gisette), cancer
classification and prediction on genetic data (Colon, Lung181, DLBCL, Prostate,
Arcene, REGED0 and MARTI0),   generic feature selection (Madelon), pharmacology
(Sido0), and image classification (PASCAL VOC 2007-2012).  We have chosen these
datasets in order to present a diverse set of challenges to  the feature
selection algorithms.  This table also reports -- to the best of our knowledge
-- the state-of-the-art (SoA) for
each dataset.

We use  linear SVM to asses the classification performance of the feature
selection algorithms. To set the parameters in our models, namely the tradeoff
parameter $\alpha$ and the $C$ parameter of the linear SVM, we used 5-fold
cross validation on training data.

\begin{table*}
    \caption{Effect of pre-processing method on unsupervised feature selection. AUC (\%) on different datasets of SVM classification, averaging the performance obtained with the first 10, 50, 100, 150, and 200 features (unsupervised feature selection).}
    \label{table:standardized_data_on_ifs}
    \centering
    %\begin{adjustbox}{width=1\textwidth}
    \small
        \begin{tabular}{ lccccccccccccccc }
          \hline
           & \multicolumn{6}{c}{Accuracy} \\
          \cmidrule(r){2-7}
          & \multicolumn{2}{c}{Original Data}& \multicolumn{2}{c}{Normalized Data}& \multicolumn{2}{c}{Standardized Data}\\
          \cmidrule(r){2-3}\cmidrule(r){4-5}\cmidrule(r){6-7}
          Dataset& avg & max & avg &  max & avg & max\\
          \hline
          Colon    & 79.79 & 82.68 & 87.12 & 90.51 & 79.85 & 89.98 \\
          USPS     & 90.81 & 95.66 & 90.66 & 95.65 & 87.70 & 91.83 \\
          Madelon  & 60.84 & 61.89 & 61.84 & 63.99 & 55.86 & 60.67 \\
          GINA     & 71.90 & 79.93 & 79.07 & 86.53 & 81.83 & 91.03 \\
          Prostate & 93.39 & 96.46 & 87.51 & 95.87 & 87.10 & 93.84 \\
          \hline
          \hline
          Mean     & 79.34 & 83.32 & \textbf{81.24} & \textbf{86.51} & 78.46 & 85.47  \\
          \hline

        \end{tabular}
    %\end{adjustbox}
\end{table*}

\subsection{Preliminary Experiments}
\label{sec:preliminary_exps}
Here, we study the effects of different  ways of constructing the adjacency matrix and different data pre-processing schemes on the classification performances of IFS and SIFS algorithms. We report the results  on five smaller datasets USPS, GINA, Colon, Prostate and Madelon.

We consider three pre-processing schemes: 1) no pre-processing (i.e. original
data), 2) standardization where each feature is transformed to zero mean and
unit variance, and 3) normalization where each  feature is transformed into the
interval $[0,1]$.
%\begin{equation}
%=\frac{k-f_{min}}{f_{max}-f_{min}}
%\end{equation}
The standard deviation constitutes an important part of the pairwise energy term
in generating the adjacency matrix in IFS algorithm \cite{Roffo15}. Table
\ref{table:standardized_data_on_ifs} reports the effects of the three data
pre-processing schemes on the
classification accuracy of IFS.

The classification accuracy is reported in two ways: avg and max.  First, the
feature selector ranks all the features. Then, a linear SVM is trained and
tested using the top $N$  features, yielding classification accuracy (percent
correct). Considering all such accuracies obtained for $N \in
\{10,50,100,150,200 \}$,  ``avg'' refers to the average of them  and ``max''
refers to the maximum.  ``avg'' has been used by \cite{Roffo15}, so do we in
order to be compatible, however, we also report ``max'' in all our experiments. 

Considering the pre-processing methods, ``normalization" yields better
classification performance (than ``no-preprocessing'') for IFS \ref{table:standardized_data_on_ifs}. However, standardization has a reverse effect, except for GINA. When using
standardized data, all the features have the same standard deviation 1, and
therefore, we expect smaller (near 0) $\alpha$ values, representing more
importance of the Spearman’s correlation coefficient part. However, our
experimental results are incompatible  with this expectation. For all the five datasets,
the returned best $\alpha$ value is 1. This means that the IFS algorithm does
not really use SPR and ranks the features based on their order in the dataset.
Moreover, these results show that the Spearman's correlation coefficient alone is not a good feature ranking method and it should be used in combination with other measures.

\begin{table*}
\footnotesize
    \caption{Effects of redundancy (as measured by Spearman's rank correlation
    (SPR) or mutual information based redundancy (RDN)) and data pre-processing
    method, without using a  relevance measure,  on unsupervised feature selection.}
\label{table:rdn_vs_spr}
\centering
\begin{adjustbox}{width=1\textwidth}
\small
\begin{tabular}{ lcccc|cccc|cccc }
  \hline
  & \multicolumn{12}{c}{Accuracy} \\
  \cmidrule(r){2-13}
  & \multicolumn{4}{c}{Original Data}& \multicolumn{4}{c}{Normalized Data}& \multicolumn{4}{c}{Standardized Data}\\
  \cmidrule(r){2-5}\cmidrule(r){6-9}\cmidrule(r){10-13}
  & \multicolumn{2}{c}{SPR}& \multicolumn{2}{c}{RDN}& \multicolumn{2}{c}{SPR} & \multicolumn{2}{c}{RDN} & \multicolumn{2}{c}{SPR} & \multicolumn{2}{c}{RDN}\\
  \cmidrule(r){2-3}\cmidrule(r){4-5}\cmidrule(r){6-7}\cmidrule(r){8-9}\cmidrule(r){10-11}\cmidrule(r){12-13}
 Dataset  & avg & max & avg &  max & avg & max & avg & max & avg &  max & avg & max\\
  \hline
  Colon	  &59.92 & 75.90    & 79.91 & 87.16 & 65.46 & 80.48 & 81.34 & 86.21 & 58.06 & 67.22 & 80.48 & 84.14\\
  USPS	  &84.89 & 93.74	& 86.75 & 95.69	& 85.11 & 94.84	& 88.84 & 95.91	& 83.31 & 91.44	& 83.42 & 91.05\\
  Madelon &50.06 & 51.19	& 51.89 & 55.94	& 50.48 & 52.28	& 57.04 & 60.37	& 49.74 & 50.70	& 50.61 & 51.14\\
  GINA	  &62.31 & 75.99	& 63.30 & 71.45	& 66.45 & 80.79	& 64.47 & 76.96	& 66.50 & 79.69	& 73.92 & 85.90\\
  Prostate&77.99 & 88.95	& 89.42 & 97.76	& 80.41 & 91.32	& 94.13 & 96.25	& 80.39 & 92.51	& 92.06 & 97.50\\

  \hline
  \hline
  Mean    & 67.03  &  77.15	& \textbf{74.25} & \textbf{81.60} &	69.58 & 79.94 &	\textbf{77.16} &  \textbf{83.14}	& 67.60  & 76.31 & \textbf{76.09} &  \textbf{81.94}  \\

  \hline

\end{tabular}
\end{adjustbox}
\end{table*}

\begin{table*}
\footnotesize
\caption{
    Effects of redundancy (as measured by SPR or RDN) and data pre-processing
    method, together with standard-deviation (STD) based relevance, on
    unsupervised     feature selection.
}
\label{table:rdn_vs_spr_on_ifs}
\centering
\begin{adjustbox}{width=1\textwidth}
\small
\begin{tabular}{ lcccc|cccc|cccc }
  \hline
  & \multicolumn{12}{c}{Accuracy} \\
  \cmidrule(r){2-13}
  & \multicolumn{4}{c}{Original Data}& \multicolumn{4}{c}{Normalized Data}& \multicolumn{4}{c}{Standardized Data}\\
  \cmidrule(r){2-5}\cmidrule(r){6-9}\cmidrule(r){10-13}
  & \multicolumn{2}{c}{SPR}& \multicolumn{2}{c}{RDN}& \multicolumn{2}{c}{SPR} & \multicolumn{2}{c}{RDN} & \multicolumn{2}{c}{SPR} & \multicolumn{2}{c}{RDN}\\
  \cmidrule(r){2-3}\cmidrule(r){4-5}\cmidrule(r){6-7}\cmidrule(r){8-9}\cmidrule(r){10-11}\cmidrule(r){12-13}
 Dataset  & avg & max & avg &  max & avg & max & avg & max & avg &  max & avg & max\\
  \hline
  Colon	    &  79.79   &  82.68	& 85.54  &  91.97 &	87.12   &  90.51 & 88.51 & 89.88	& 79.85   &  89.98	& 88.86 & 91.46\\
  USPS	    &  90.81   &  95.66	& 90.60  &  95.64 & 90.66   &  95.65 & 90.86 & 95.89	& 87.70   &  91.83	& 87.87 & 92.78\\
  Madelon	&  60.84   &  61.89 & 61.46  &  62.46 &	61.84   &  63.99 & 62.16 & 63.91	& 55.86   &  60.67	& 53.50 & 57.94\\
  GINA	    &  71.90   &  79.93	& 70.40  &  80.39 & 79.07   &  86.53 & 79.58 & 87.09	& 81.83   &  91.03	& 79.56 & 89.50\\
  Prostate	&  93.39   &  96.46	& 94.28  &  98.16 & 87.51   &  95.87 & 94.67 & 98.02	& 87.10   &  93.84	& 93.11 & 97.51\\

  \hline
  \hline
  Mean      &  79.34   &  83.32	& \textbf{80.45}  &  \textbf{85.72} & 81.24   &  86.51 & \textbf{83.15} & \textbf{86.95}    & 78.46   &  85.47  & \textbf{80.58} & \textbf{85.83} \\

  \hline

\end{tabular}
\end{adjustbox}
\end{table*}

Table \ref{table:rdn_vs_spr} reports the effects of using SPR or RDN -- which
are two different choices to measure redundancy --  alone in
the construction of  the adjacency matrix. As it can be seen, RDN is superior in
most of the cases and shows increases of up to $8\%$ for all the three data
formats. SPR is not able to individuate non-monotonic dependencies between
features and therefore more complex functional dependencies between features are
not measured. On the other hand, RDN uses mutual information, which is able to
individuate any kind of dependency (linear and non-linear) between features.
Table \ref{table:rdn_vs_spr_on_ifs} reports the effects of using SPR/RDN when
they are used in combination with standard deviation (STD)  based relevance. As
it can be seen, RDN is superior again for this adjacency matrix setting. When we
use the mutual information based relevance  (Table
\ref{table:rdn_vs_spr_on_sifs}), the results are slightly different and the SPR
shows better classification performance. Moreover, we get the best
classification performance for standardized data format, which is in contrast with the unsupervised matrix settings.

In summary, the following two important results can be derived from all these
preliminary experiments:
\begin{enumerate}
  \item  For \emph{unsupervised} feature selection, \emph{normalizing} the data
      and then using \emph{STD based relevance} in combination with \emph{RDN
        based redundancy} yield the best classification performance. This
        corresponds to our `modified infinite feature selection' method , mIFS.
  \item For \emph{supervised} feature selection, \emph{standardizing} the data
      and then using \emph{mutual information based relevance} in combination
        with \emph{SPR based redundancy} gives the  best classification
        performance. This corresponds to our `supervised infinite feature
        selection' method, SIFS.
\end{enumerate}

\begin{table*}
\footnotesize
\caption{
    Effects of redundancy (as measured by SPR or RDN) and data pre-processing
    method, together with mutual information (MI) based relevance, on
    supervised feature selection
}
\label{table:rdn_vs_spr_on_sifs}
\centering
\begin{adjustbox}{width=1\textwidth}
\small
\begin{tabular}{ lcccc|cccc|cccc }
\hline
& \multicolumn{12}{c}{Accuracy} \\
\cmidrule(r){2-13}
& \multicolumn{4}{c}{Original Data}& \multicolumn{4}{c}{Normalized Data}& \multicolumn{4}{c}{Standardized Data}\\
\cmidrule(r){2-5}\cmidrule(r){6-9}\cmidrule(r){10-13}
& \multicolumn{2}{c}{SPR}& \multicolumn{2}{c}{RDN}& \multicolumn{2}{c}{SPR} & \multicolumn{2}{c}{RDN} & \multicolumn{2}{c}{SPR} & \multicolumn{2}{c}{RDN}\\
\cmidrule(r){2-3}\cmidrule(r){4-5}\cmidrule(r){6-7}\cmidrule(r){8-9}\cmidrule(r){10-11}\cmidrule(r){12-13}
Dataset  & avg & max & avg &  max & avg & max & avg & max & avg &  max & avg & max\\
\hline
Colon	   & 93.31   &  97.00	& 93.07 & 97.85 & 90.46  &  92.14   & 91.98 & 95.66	& 94.67  &  97.71	& 93.06 & 95.79\\
USPS	   & 92.18   &  95.64	& 92.28 & 96.06	& 90.06  &  95.71	& 91.39 & 95.89 & 89.42  & 93.21	& 89.33 & 93.05\\
Madelon  & 60.93   &  62.78	& 62.00 & 62.41	& 61.63  &  62.65	& 59.16 & 60.89	& 63.83  & 64.46	& 63.55 & 63.97\\
GINA	   & 86.17   &  89.22	& 82.08 & 88.80	& 91.30  &  93.08	& 91.01 & 92.93 & 93.09  & 92.74	& 90.94 & 93.09\\
Prostate & 97.85   &  98.39	& 97.71 & 98.80	& 96.24  &  98.57	& 93.12 & 96.38	& 98.31  & 98.84	& 98.09 & 98.78\\

\hline
\hline
Mean     & \textbf{86.08} &  \textbf{88.60}	& 85.42 &  88.78	&\textbf{85.93 }&  \textbf{88.43}	& 85.33 &  88.35	& \textbf{87.86} &  \textbf{89.39}	& 86.99  &  88.93\\

\hline

\end{tabular}
\end{adjustbox}
\end{table*}

\begin{table*}
\caption{
    Classification accuracies obtained using different feature selectors, namely
    IFS \cite{Roffo15}, mRMR \cite{Peng05}, mIFS (ours) and SIFS (ours).  (See
    Section \ref{sec:preliminary_exps} for the explanations of  ``avg'' and ``max''.)
}
\label{table:feature_selection}
\centering
\footnotesize
\begin{adjustbox}{width=1\textwidth}
\begin{tabular}{lcccc||cccc}

\hline
%\multicolumn{2}{c}{Item} \\
%\cline{1-2}
& \multicolumn{2}{c}{IFS}& \multicolumn{2}{c}{mIFS}&
\multicolumn{2}{c}{mRMR}& \multicolumn{2}{c}{SIFS} \\
\cmidrule(r){2-3}\cmidrule(r){4-5}\cmidrule(r){6-7}\cmidrule(r){8-9}
Dataset  & avg & max & avg &  max & avg & max & avg & max \\
\hline
USPS     &  90.66 & 95.65  &     \textbf{90.86} & \textbf{95.89}      &  91.11 & 93.28  & 89.42 & 93.21   \\
GINA     &  79.07 & 86.53  &    \textbf{79.58}  & \textbf{87.09}    &   91.98 & 92.86 &   \textbf{92.74} & 93.09  \\
Gissete  & 95.94 & 97.62   &    95.93 & 97.62    &   97.75 & 99.06  &   96.66 & 98.64   \\
Colon    &    87.12 & 90.51    &       \textbf{88.51}& 89.88        &   89.03 & 91.32 &     \textbf{94.67} & \textbf{97.71}   \\
Lung181  & 99.14 & 100.00    & \textbf{99.51} & 100.00    &   99.87 & 100.00 & \textbf{100.00}    &    \textbf{100.00}  \\
DLBCL    &  99.50 & 100.00      & \textbf{99.63} & 100.00    & 96.90 & 99.23   & \textbf{99.10} & \textbf{100.00}    \\
Prostate &  87.51 & 95.87    &    \textbf{94.67} & \textbf{98.02}    & 97.25 & 97.84  &  \textbf{98.31} & \textbf{98.84}   \\
Arcene   &  74.09 & 82.18     & \textbf{86.23} & \textbf{88.55 }     & 76.35 & 83.28  &  \textbf{80.12} & 82.67     \\
REGED0   &  81.98 & 95.57     &   \textbf{83.86} & \textbf{95.92}    & 99.13 & 99.79   & \textbf{99.70} & \textbf{99.87}  \\
MARTI0   & 65.98 & 73.16     &  59.34 & 72.87     & 79.31 & 90.41  &   \textbf{83.70} & \textbf{91.23}  \\
Madelon  &  59.54 & 61.79     &  \textbf{ 61.00} & \textbf{62.55} &  58.82 & 61.13  & \textbf{60.70} & \textbf{63.03}\\
Sido0    &  \textbf{87.07} & \textbf{91.98}    &  86.96 & 91.88   &  87.20 & 91.13    & \textbf{92.26} & \textbf{92.80}  \\
\\
\hline
\hline
Average     &  83.97 & 89.23    &  \textbf{85.75} & \textbf{90.02}   &  88.72 & 91.13   & \textbf{90.61} & \textbf{92.59}   \\

\hline
\end{tabular}
\end{adjustbox}
\end{table*}

\subsection{Comparison with IFS and mRMR}
Here, we compare the performances of the proposed infinite feature selection
algorithms with the state-of-the-art algorithms. For unsupervised feature
selection, we compare the original  IFS method  \cite{Roffo15} with mIFS, our
proposed method for unsupervised problems. All features are normalized before
feature selection. On 9 out of 12 datasets mIFS outperforms IFS (Table
\ref{table:feature_selection}. Specifically,  we report  12\% improvement on
\emph{Arcene} and 7\% on \emph{Prostate} datasets.

For supervised feature selection, we compare mRMR \cite{Peng05} -- arguably,
the most well known information theoretic feature selection algorithm -- with SIFS, our
proposed method for supervised problems. All features are standardized before
feature selection. On 10 out of 12 datasets, SIFS outperforms mRMR (Table
\ref{table:feature_selection}) with an   average improvement of 1.89\% in
classification accuracy.

Finally, we compare IFS \cite{Roffo15} with our SIFS. On average (over 12
datasets), SIFS outperforms IFS with a margin of about 6\% in classification
accuracy, which shows the impact of using supervision for feature selection.

\subsection{ Image classification experiments on PASCAL VOC datasets}

The
experiments here considers a  combination of feature selection and linear SVM
applied to convolutional neural network (CNN) based features.  We extracted CNN
features from the penultimate layers of the ResNet \cite{He15} (1000 features),
GoogleNet \cite{Szegedy15} (1000 features), and
VGG-VD \cite{simonyan14} (4096 features) deep networks. We used the models, pre-trained on
ILSVRC, from the MatConvNet distribution \cite{vedaldi15}. On each of the three
feature sets,
we applied normalization and our supervised infinite feature selection. Then, we
trained  a linear SVM per set and averaged the three SVM scores to obtain final classification scores.  Table \ref{table:voc_results} shows the mean average-precision (mAP) results for the
PASCAL VOC 2007 and 2012 datasets, using different feature selectors.
Our method, SIFS, outperforms both IFS and mRMR on both datasets\footnote{Leaderboard
snapshot taken in December 2016:\\
\url{http://user.ceng.metu.edu.tr/~emre/resources/SIFS_PASCAL_result.png}. Our
submission is named "SE."
Anonymous results link:
\url{http://host.robots.ox.ac.uk:8080/anonymous/MV5IFE.html}.  Live
leaderboard:
\url{http://host.robots.ox.ac.uk:8080/leaderboard/displaylb.php?challengeid=11&compid=1}}.

\begin{table*}
\footnotesize
\caption{
    mAP (\%) results obtained using  different feature selectors on the PASCAL VOC object recognition datasets. The numbers in parentheses are the percentages of features kept by the approach after the cross-validation phase for ResNet, GoogleNet and VGG-VD, respectively. }
\label{table:voc_results}
\centering
%\begin{adjustbox}{width=1\textwidth}
\small
\begin{tabular}{ lcccc }
\hline
Dataset & SoA & No feature selection & mRMR & SIFS \\
\hline
VOC 2007 & 83.5\% & 84.63\% & 84.95\% & \textbf{85.90}\%\\
       &        & &(60,70,70) & (40,60,60)\\
       \hline
VOC 2012 & 85.4\% & 85.78\% & 85.88\% & \textbf{86.50}\%\\
        &        & &(50,70,70) & (40,60,60)\\

\hline
\end{tabular}
%\end{adjustbox}
\end{table*}

\section{Conclusions } \label{section:Conclusions}
In this paper we present two new ways of constructing the feature adjacency
matrix for the infinite feature selection method. For unsupervised feature
selection, we propose the mIFS method which   uses a combination of standard-deviation based relevance and mutual
information based redundancy. For supervised feature selection, we propose the
SIFS method which uses  a combination of mutual information based relevance and spearman's
rank correlation based redundancy. We tested the accuracy of the proposed
methods on $14$ high dimensional benchmark datasets using linear SVM. Our
proposed methods, mIFS and SIFS, gave top performances on most of the benchmark
datasets beating both IFS \cite{Roffo15} and mRMR \cite{Peng05}. 
%On five datasets, SIFS produced the new
%state of the art, including the two PASCAL VOC datasets.
Our source code is available at GitHub\footnote{
\url{https://github.com/Sadegh28/SIFS}}  for the
sake of reproducibility of our results.

\bibliographystyle{acm}
\bibliography{References2}

\begin{thebibliography}{10}

\bibitem{Alon99}
{\sc Alon, U., Barkai, N., Notterman, D.~A., Gish, K., Ybarra, S., Mack, D.,
  and Levine, A.~J.}
\newblock Broad patterns of gene expression revealed by clustering analysis of
  tumor and normal colon tissues probed by oligonucleotide arrays.
\newblock In {\em Proceedings of the National Academy of Sciences\/} (1999),
  pp.~6745--6750.

\bibitem{Battiti94}
{\sc Battiti, R.}
\newblock Using mutual information for selecting features in supervised neural
  net learning.
\newblock {\em IEEE Transactions on Neural Networks 5}, 4 (1994), 537--550.

\bibitem{Canedo13}
{\sc Bol{\'o}n-Canedo, V., S{\'a}nchez-Maro{\~{n}}o, N., and Alonso-Betanzos,
  A.}
\newblock A review of feature selection methods on synthetic data.
\newblock {\em Knowledge and Information Systems 34}, 3 (2013), 483--519.

\bibitem{Brown12}
{\sc Brown, G., Pocock, A., Zhao, M.-J., and Luj\'{a}n, M.}
\newblock Conditional likelihood maximisation: A unifying framework for
  information theoretic feature selection.
\newblock {\em Journal of Machine Learning Research 13\/} (2012), 27--66.

\bibitem{Cawley08}
{\sc Cawley, G.~C.}
\newblock Causal and non-causal feature selection for ridge regression.
\newblock In {\em WCCI Causation and Prediction Challenge\/} (2008),
  pp.~107--128.

\bibitem{Chang08}
{\sc Chang, Y.-W., and Lin, C.-J.}
\newblock Feature ranking using linear svm.
\newblock In {\em WCCI Causation and Prediction Challenge\/} (2008),
  pp.~53--64.

\bibitem{Chapelle06}
{\sc Chapelle, O., Schölkopf, B., and Zien, A.}
\newblock {\em Semi\-Supervised Learning}.
\newblock MIT press, 2006.

\bibitem{Nips03}
{\sc Clopinet}.
\newblock {Feature Selection Challenge},{ NIPS 2003}.
\newblock http://clopinet.com/isabelle/Projects/NIPS2003/, 2003.
\newblock [Online; accessed 06-March-2015].

\bibitem{WCCI06}
{\sc Clopinet}.
\newblock {Performance Prediction Challenge},{ WCCI 2006}.
\newblock http://clopinet.com/isabelle/Projects/modelselect/, 2006.
\newblock [Online; accessed 06-March-2015].

\bibitem{WCCI08}
{\sc Clopinet}.
\newblock {Causation and Prediction Challenge},{ WCCI 2008}.
\newblock http://www.causality.inf.ethz.ch, 2008.
\newblock [Online; accessed 06-March-2015].

\bibitem{Dash03}
{\sc Dash, M., and Liu, H.}
\newblock Consistency-based search in feature selection.
\newblock {\em Artificial Intelligence 151}, 1 (2003), 155--176.

\bibitem{Diaz06}
{\sc D{\'\i}az-Uriarte, R., and De~Andres, S.~A.}
\newblock Gene selection and classification of microarray data using random
  forest.
\newblock {\em BMC bioinformatics 7}, 1 (2006), 1.

\bibitem{Dionisio04}
{\sc Dionisio, A., Menezes, R., and Mendes, D.~A.}
\newblock Mutual information: a measure of dependency for nonlinear time
  series.
\newblock {\em Physica A: Statistical Mechanics and its Applications 344}, 1
  (2004), 326--329.

\bibitem{Eskandari16a}
{\sc Eskandari, S., and Javidi, M.~M.}
\newblock Online streaming feature selection using rough sets.
\newblock {\em International Journal of Approximate Reasoning 69}, C (2016),
  35--57.

\bibitem{pascal-voc-2007}
{\sc Everingham, M., Van~Gool, L., Williams, C. K.~I., Winn, J., and Zisserman,
  A.}
\newblock The {PASCAL} {V}isual {O}bject {C}lasses {C}hallenge 2007 {(VOC2007)}
  {R}esults.
\newblock
  http://www.pascal-network.org/challenges/VOC/voc2007/workshop/index.html,
  2007.

\bibitem{pascal-voc-2012}
{\sc Everingham, M., Van~Gool, L., Williams, C. K.~I., Winn, J., and Zisserman,
  A.}
\newblock The {PASCAL} {V}isual {O}bject {C}lasses {C}hallenge 2012 {(VOC2012)}
  {R}esults.
\newblock
  http://www.pascal-network.org/challenges/VOC/voc2012/workshop/index.html,
  2012.

\bibitem{Fleuret04}
{\sc Fleuret, F.}
\newblock Fast binary feature selection with conditional mutual information.
\newblock {\em Journal of Machine Learning Research 5\/} (2004), 1531--1555.

\bibitem{Gordon02}
{\sc Gordon, G.~J., Jensen, R.~V., Hsiao, L.-L., Gullans, S.~R., Blumenstock,
  J.~E., Ramaswamy, S., Richards, W.~G., Sugarbaker, D.~J., and Bueno, R.}
\newblock Translation of microarray data into clinically relevant cancer
  diagnostic tests using gene expression ratios in lung cancer and
  mesothelioma.
\newblock {\em Cancer Research 62}, 17 (2002), 4963--4967.

\bibitem{Guyon08}
{\sc Guyon, I., Aliferis, C.~F., Cooper, G.~F., Elisseeff, A., Pellet, J.-P.,
  Spirtes, P., and Statnikov, A.~R.}
\newblock Design and analysis of the causation and prediction challenge.
\newblock In {\em WCCI Causation and Prediction Challenge\/} (2008), pp.~1--33.

\bibitem{Guyon03}
{\sc Guyon, I., and Elliseff, A.}
\newblock An {I}ntroduction to {V}ariable and {F}eature {S}election.
\newblock {\em {J}ournal of {M}achine {L}earning {R}esearch 3\/} (2003),
  1157--1182.

\bibitem{Guyon07}
{\sc Guyon, I., Li, J., Mader, T., Pletscher, P.~A., Schneider, G., and Uhr,
  M.}
\newblock Competitive baseline methods set new standards for the \{NIPS\} 2003
  feature selection benchmark.
\newblock {\em Pattern Recognition Letters 28}, 12 (2007), 1438 -- 1444.

\bibitem{Guyon08b}
{\sc Guyon, I., Saffari, A., Dror, G., and Cawley, G.}
\newblock Analysis of the \{IJCNN\} 2007 agnostic learning vs. prior knowledge
  challenge.
\newblock {\em Neural Networks 21}, 2–3 (2008), 544 -- 550.

\bibitem{He15}
{\sc He, K., Zhang, X., Ren, S., and Sun, J.}
\newblock Deep residual learning for image recognition.
\newblock {\em arXiv preprint arXiv:1512.03385\/} (2015).

\bibitem{Kohavi97}
{\sc Kohavi, R., and John, G.~H.}
\newblock Wrappers for feature subset selection.
\newblock {\em Artificial Intelligence 97}, 1–2 (1997), 273--324.

\bibitem{Koller96}
{\sc Koller, D., and Sahami, M.}
\newblock Toward optimal feature selection.
\newblock Tech. Rep. 1996-77, Stanford InfoLab, 1996.

\bibitem{Lewis92}
{\sc Lewis, D.~D.}
\newblock Feature selection and feature extraction for text categorization.
\newblock In {\em Proceedings of the Workshop on Speech and Natural Language\/}
  (Stroudsburg, PA, USA, 1992), HLT '91, pp.~212--217.

\bibitem{Lovato12}
{\sc Lovato, P., Bicego, M., Cristani, M., Jojic, N., and Perina, A.}
\newblock Feature selection using counting grids: Application to microarray
  data.
\newblock In {\em Structural, Syntactic, and Statistical Pattern Recognition:
  Joint IAPR International Workshop, SSPR{\&}SPR 2012, Hiroshima, Japan,
  November 7-9, 2012. Proceedings\/} (2012), G.~Gimel'farb, E.~Hancock,
  A.~Imiya, A.~Kuijper, M.~Kudo, S.~Omachi, T.~Windeatt, and K.~Yamada, Eds.,
  pp.~629--637.

\bibitem{Maji09}
{\sc Maji, S., and Malik, J.}
\newblock Fast and accurate digit classification.
\newblock {\em EECS Department, University of California, Berkeley, Tech. Rep.
  UCB/EECS-2009-159\/} (2009).

\bibitem{Meyer06}
{\sc Meyer, P.~E., and Bontempi, G.}
\newblock {\em On the Use of Variable Complementarity for Feature Selection in
  Cancer Classification}.
\newblock Springer Berlin Heidelberg, Berlin, Heidelberg, 2006, pp.~91--102.

\bibitem{Neal06}
{\sc Neal, R.~M., and Zhang, J.}
\newblock {\em High Dimensional Classification with Bayesian Neural Networks
  and Dirichlet Diffusion Trees}.
\newblock Springer Berlin Heidelberg, Berlin, Heidelberg, 2006, pp.~265--296.

\bibitem{Peng05}
{\sc Peng, H., Long, F., and Ding, C.}
\newblock Feature selection based on mutual information criteria of
  max-dependency, max-relevance, and min-redundancy.
\newblock {\em IEEE Transactions on Pattern Analysis and Machine Intelligence
  27}, 8 (2005), 1226--1238.

\bibitem{Roffo15}
{\sc Roffo, G., Melzi, S., and Cristani, M.}
\newblock Infinite feature selection.
\newblock In {\em The IEEE International Conference on Computer Vision
  (ICCV)\/} (December 2015).

\bibitem{Saeys07}
{\sc Saeys, Y., Inza, I., and Larra{\~{n}}aga, P.}
\newblock A review of feature selection techniques in bioinformatics.
\newblock {\em Bioinformatics 23}, 19 (2007), 2507--2517.

\bibitem{Shipp02}
{\sc Shipp, M.~A., Ross, K.~N., Tamayo, P., Weng, A.~P., Kutok, J.~L., Aguiar,
  R.~C., Gaasenbeek, M., Angelo, M., Reich, M., Pinkus, G.~S., et~al.}
\newblock Diffuse large b-cell lymphoma outcome prediction by gene-expression
  profiling and supervised machine learning.
\newblock {\em Nature medicine 8}, 1 (2002), 68--74.

\bibitem{simonyan14}
{\sc Simonyan, K., and Zisserman, A.}
\newblock Very deep convolutional networks for large-scale image recognition.
\newblock {\em arXiv preprint arXiv:1409.1556\/} (2014).

\bibitem{Singh02}
{\sc Singh, D., Febbo, P.~G., Ross, K., Jackson, D.~G., Manola, J., Ladd, C.,
  Tamayo, P., Renshaw, A.~A., D'Amico, A.~V., Richie, J.~P., et~al.}
\newblock Gene expression correlates of clinical prostate cancer behavior.
\newblock {\em Cancer cell 1}, 2 (2002), 203--209.

\bibitem{Szegedy15}
{\sc Szegedy, C., Liu, W., Jia, Y., Sermanet, P., Reed, S., Anguelov, D.,
  Erhan, D., Vanhoucke, V., and Rabinovich, A.}
\newblock Going deeper with convolutions.
\newblock In {\em The IEEE Conference on Computer Vision and Pattern
  Recognition (CVPR)\/} (June 2015).

\bibitem{vedaldi15}
{\sc Vedaldi, A., and Lenc, K.}
\newblock Matconvnet -- convolutional neural networks for matlab.
\newblock In {\em Proceeding of the {ACM} Int. Conf. on Multimedia\/} (2015).

\bibitem{Wu13}
{\sc Wu, X., Yu, K., Ding, W., Wang, H., and Zhu, X.}
\newblock Online feature selection with streaming features.
\newblock {\em IEEE Transactions on Pattern Analysis and Machine Intelligence
  35}, 5 (2013), 1178--1192.

\bibitem{Yu04}
{\sc Yu, L., and Liu, H.}
\newblock Efficient feature selection via analysis of relevance and redundancy.
\newblock {\em Journal of Machine Learning Research 5\/} (2004), 1205--1224.

\end{thebibliography}

\end{document}